%% file: main.tex
\documentclass[10pt,twocolumn,letterpaper]{article}

\usepackage[pagenumbers]{wacv} 

\usepackage{times}
\usepackage{epsfig}
\usepackage{graphicx}
\usepackage{amsmath}
\usepackage{amssymb}
\usepackage{booktabs}
\usepackage{multirow}
\usepackage{colortbl}
\usepackage{algorithm}
\usepackage{algorithmic}
\usepackage{lipsum}
\usepackage{arydshln}
\usepackage{pifont}
\usepackage{color, colortbl}
\usepackage{breqn}
\usepackage{comment}
\usepackage[accsupp]{axessibility} 
\usepackage[capitalize]{cleveref}
\crefname{section}{Sec.}{Secs.}
\Crefname{section}{Section}{Sections}
\Crefname{table}{Table}{Tables}
\crefname{table}{Tab.}{Tabs.}

\usepackage{array}
\usepackage{stfloats}
\newcommand{\xmark}{\ding{55}}
\newcommand{\cmark}{\ding{51}}
\usepackage[table,x11names]{xcolor}
\definecolor{lightgray}{gray}{0.9}

\def\wacvPaperID{00657} 
\def\confName{WACV}
\def\confYear{2025}

\begin{document}
\title{CLFace: A Scalable and Resource-Efficient Continual Learning Framework for Lifelong Face Recognition}

\author{Md Mahedi Hasan, Shoaib Meraj Sami, and Nasser Nasrabadi\\
	West Virginia University, Morgantown, West Virginia, USA\\
	{\tt\small mh00062@mix.wvu.edu, sms00052@mix.wvu.edu, nasser.nasrabadi@mail.wvu.edu}
}

\maketitle

\begin{abstract}
An important aspect of deploying face recognition (FR) algorithms in real-world applications is their ability to learn new face identities from a continuous data stream. However, the online training of existing deep neural network-based FR algorithms, which are pre-trained offline on large-scale stationary datasets, encounter two major challenges: (I) catastrophic forgetting of previously learned identities, and (II) the need to store past data for complete retraining from scratch, leading to significant storage constraints and privacy concerns. In this paper, we introduce CLFace, a continual learning framework designed to preserve and incrementally extend the learned knowledge. CLFace eliminates the classification layer, resulting in a resource-efficient FR model that remains fixed throughout lifelong learning and provides label-free supervision to a student model, making it suitable for open-set face recognition during incremental steps. We introduce an objective function that employs feature-level distillation to reduce drift between feature maps of the student and teacher models across multiple stages. Additionally, it incorporates a geometry-preserving distillation scheme to maintain the orientation of the teacher model's feature embedding. Furthermore, a contrastive knowledge distillation is incorporated to continually enhance the discriminative power of the feature representation by matching similarities between new identities. Experiments on several benchmark FR datasets demonstrate that CLFace outperforms baseline approaches and state-of-the-art methods on unseen identities using both in-domain and out-of-domain datasets.
\end{abstract}

\section{Introduction} ~\label{sec:intro}
Despite remarkable progress in face recognition (FR) in recent years~\cite{arcface,magface,adaface}, several critical areas still require attention to make FR models viable for real-world surveillance applications. The primary challenge in real-world deployment is updating existing FR models when batches of new identities emerge continuously in data streams, which may contain a mix of new and previously learned identities. For example, an FR model deployed in a smart surveillance system at an airport must continually learn to identify thousands of new identities each day. However, state-of-the-art deep neural network-based FR models, typically trained offline using stationary datasets, cannot adapt dynamically without requiring complete retraining. Moreover, retraining these networks on large-scale datasets presents practical challenges, including high computational demands, storage constraints, and privacy concerns~\cite{lwf}. 

Fine-tuning the learned model with new identities could be a potential approach. However, this often leads the model to experience \textit{catastrophic forgetting} (CF)~\cite{mccloskey89, goodfellow_13, icarl} of previously learned knowledge, as the feature space becomes biased toward the new task. Alternatively, storing or replaying old samples can also be an effective approach for incremental learning~\cite{gen_replay, icarl,podnet}. However, storing exemplars or prototypes is impractical for biometric applications like FR, which often handle millions of unique identities~\cite{crl}. As a result, continual learning (CL) techniques~\cite{podnet,icarl,lwm,lwf,ewc} have been developed, enabling the new model (student) to train over time with only new batches of identities, while effectively mitigating the risk of CF. CL algorithms often use knowledge distillation~\cite{hinton_15, lwf, lwm} to transfer feature representations from a base/teacher model trained on source domain data to guide a student model in replicating certain aspects of it, thereby preserving the learned knowledge.

In a biometric identification (BI) system, an ideal CL algorithm should possess the following properties~\cite{lwm, icarl}:
(I) it should handle continuous streams of batches with new and previously learned identities in any order;
(II) it should maintain a nearly fixed number of parameters to keep computational requirements and memory footprint bounded;
(III) it should retain past knowledge while improving its feature representation capacity incrementally; and
(IV) it should not store exemplars, prototypes, or models of previous identities. However, none of the state-of-the-art (SoTA) CL algorithms proposed in the literature meet all these criteria~\cite{podnet,icarl, ssre,lucir, zhu2021prototype}. For instance, while many of these algorithms are designed to mitigate CF, they often fail to improve performance with a new batch of classes~\cite{podnet, icarl}. Additionally, they often store memory-intensive exemplars or prototypes, posing a significant challenge for BI systems.
 
Moreover, SoTA CL algorithms typically handle a limited number of classes and employ a classification layer to optimize a general loss function, $L_{total} = \lambda_{1} L_{ID} + \lambda_{2} L_{KD}$, where $L_{ID}$ is the identity loss requiring label supervision, and $L_{KD}$ represents the knowledge distillation loss. However, in large-scale biometric applications, classifying a vast number of new identities in each incremental step is impractical, as the expanding fully connected (FC) classification layer would exceed the available memory and computational resources~\cite{crl}. Thus, conventional CL-based approaches fail to yield a scalable FR model, particularly in a resource-constrained system. To address this, we propose removing the final FC layer during incremental learning and relying solely on distillation loss within feature maps and embeddings. This approach enables us to develop a scalable model that substantially reduces computational and memory requirements for lifelong learning.

In this paper, we introduce CLFace, a continual learning framework for FR that strictly follows all the essential properties of a BI system. CLFace effectively preserves and incrementally enhances learned representations through a classification-free architecture, minimizing the risk of CF compared to label-supervised models. It offers several advantages over SoTA CL models. First, it ensures \textit{scalability} by continually learning new batches of identities using a fixed network. Second, it provides exemplar- and prototype-free solutions for lifelong FR, making it suitable for applications in systems with strict privacy and resource constraints. Third, due to its label-free supervision, it can be trained on unlabeled face datasets during incremental steps, enabling it to handle \textit{open-set face recognition}. Fourth, it tackles the \textit{concept drift} issue in FR by incrementally adapting to changes in facial attributes among known identities, ensuring continued accuracy and reliability over time.

In this study, we implement multiscale feature distillation (MSFD) loss to minimize drift between the intermediate feature maps of the student and teacher models. Additionally, we use geometry-preserving knowledge distillation (GPKD) loss to align the student model's orientation with that of the teacher model's embedding space. Furthermore, a contrastive knowledge distillation (CKD) loss is used to improve the discriminative power of the feature representation by matching similarities between new identities. These losses effectively transfer the learned feature representation to the student model, ensuring that CLFace maintains its generalizability despite the absence of identity supervision-a key for lifelong learning in BI systems.

\begin{figure}[t] 	
	\centering
	\includegraphics[width=\linewidth]{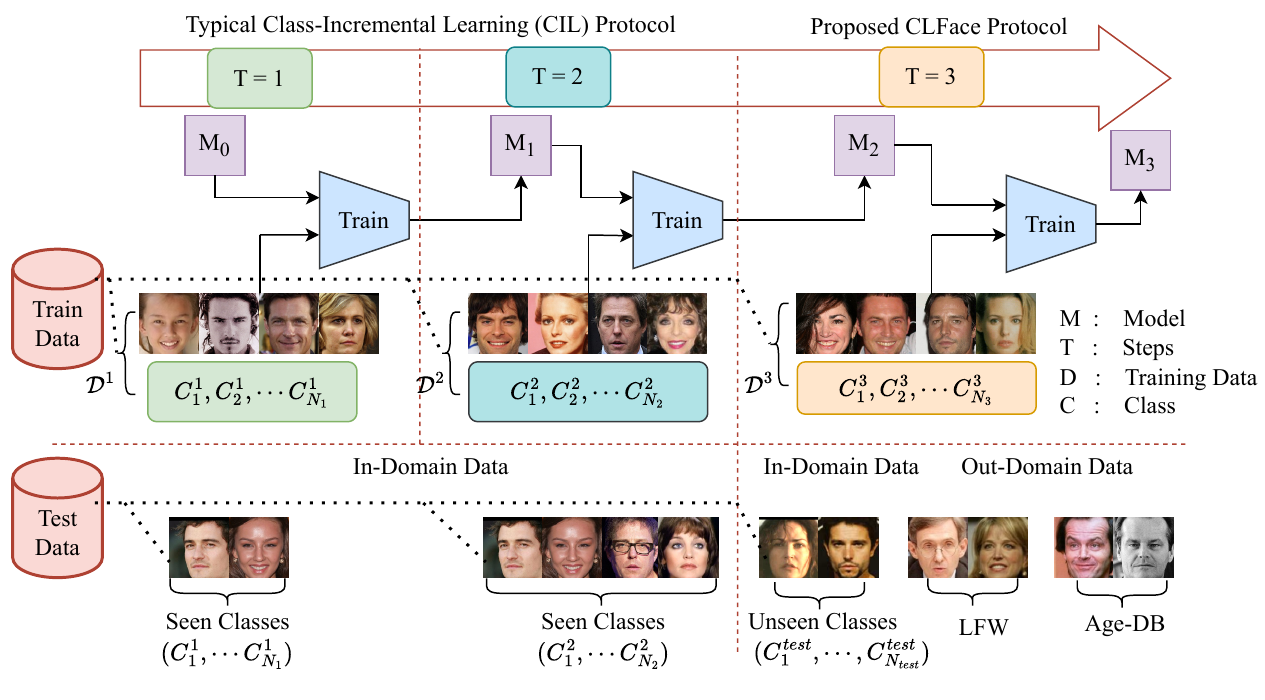}
	\caption{Overview of the training and testing protocol of the proposed CLFace algorithm. Here, we adopt a more challenging setting by testing CLFace on unseen identities from both in-domain and out-of-domain datasets, to evaluate its robustness in real-world biometric applications. This contrasts with typical CL protocols that evaluate models on seen identities within the same dataset.}
	\label{fig:cl_protocol}
	\vspace{-2mm}
\end{figure}

Typical CL protocols evaluate models on seen classes within the same dataset. This approach not only contradicts standard FR evaluation protocols~\cite{arcface, adaface} but also results in minimal CF, as the model is evaluated on the same dataset it was trained on~\cite{crl}. Therefore, we depart from evaluating CLFace on seen identities. Instead, as illustrated in Figure~\ref{fig:cl_protocol}, we evaluate CLFace on unseen identities from both in-domain and out-of-domain datasets, reflecting one of the most challenging CL scenarios. Experimental results demonstrate that our approach outperforms SoTA methods and baseline approaches across various CL scenarios. In summary, our contributions encompass three key aspects:
\begin{itemize}
	\item We propose CLFace, a scalable continual learning algorithm designed to preserve and incrementally extend learned knowledge. We remove the classification layer to develop a resource-efficient model that remains fixed throughout lifelong learning. 
	
	\vspace{-1mm}
	\item We formulate an objective function that combines MFSD and GPKD losses to enable the student model to effectively retain the learned representation of the teacher model. It also incorporates a CKD loss, that continually refines discriminative representations through similarity matching between new identities.
	
	\vspace{-1mm}
	\item We also introduce a challenging protocol that evaluates CLFace on unseen identities from both in-domain and out-of-domain datasets. Under this protocol, our CLFace outperforms SoTA methods and baseline approaches across various benchmark datasets.

\end{itemize}

\section{Related Work} \label{sec:related_work}
\paragraph{Continual learning}
Continual learning, also known as incremental learning~\cite{icarl, obj_incr, chaudhry2018, liu2020mnemonics}, has found widespread applications across various domains, including image classification~\cite{icarl,lwm,lwf,crl}, image generation~\cite{wu2018memory}, object detection~\cite{obj_incr}, and re-identification~\cite{pu2021lifelong}, among others. However, many of these algorithms still encounter the issue of CF to some extent~\cite{goodfellow_13}. Literature on mitigating CF broadly categorizes CL algorithms into four groups: (I) regularization techniques~\cite{ewc, lucir}, (II) dynamic network modifications~\cite{progress}, (III) storing exemplars~\cite{icarl, podnet}, and (IV) generative modeling~\cite{gen_replay}.
iCaRL~\cite{icarl} and its variants~\cite{podnet, lucir} employ an exemplar-based CL method including a herding technique for exemplar selection, along with a feature extractor, a nearest-mean classifier, and an objective function that combines classification and distillation losses. However, as discussed in Section~\ref{sec:intro}, storing or replaying exemplars, or dynamically expanding the model at each incremental step is impractical for most BI systems. Therefore, we prefer regularization-based CL methods, which add a penalty term to prevent changes in the mapping function~\cite{ewc}. For instance, elastic weight consolidation~\cite{ewc} reduces CF by limiting updates to important weights. Additionally, these methods utilize knowledge distillation loss to minimize feature drift from previous steps~\cite{lwf}. Learning without Forgetting (LwF)~\cite{lwf} introduces a task-incremental approach that combines knowledge distillation to retain prior knowledge from past steps, without storing any exemplars. Next, a class incremental variant of LwF, known as LwF-MC, is introduced in~\cite{icarl}.
\vspace{-2mm}
\paragraph{Face recognition}
In recent years, remarkable advancements have been made in the field of face recognition, primarily driven by the adoption of margin-based losses~\cite{arcface,magface,adaface}. However, limited attention has been given to learning new face identities from continuous data streams. While research on continual facial expression recognition~\cite{em_rec_22, nikhil_2020, nikhil_2022} and lifelong person re-ID~\cite{martinel2016temporal, pu2021lifelong, ge_lifelong_22} exists, these methods rely on storing exemplars, which differs from our approach. In a related work, Lopez et al.~\cite{lopez2022incremental} propose an online incremental learning approach for open-set FR from video surveillance, which concurrently predicts and updates an ensemble of SVM classifiers using a self-training paradigm. 

Authors in~\cite{crl} propose a scalable CL method that uses continual representation learning (CRL) for large-scale biometric applications such as FR and person Re-ID. They employ a flexible knowledge distillation scheme with techniques including neighborhood selection and consistency relaxation, to save computational resources. However, CRL~\cite{crl} relies on label supervision through a classification layer, which may limit scalability in lifelong FR that involves continuous incremental updates with a large number of identities, making it impractical in memory-constrained scenarios. Additionally, it shows some overfitting to new data, leading to poorer performance on out-of-domain datasets compared to in-domain datasets.
\vspace{-2mm}
\paragraph{Knowledge distillation}
Knowledge Distillation (KD) aims to transfer knowledge from a large pre-trained model (teacher) to a simpler, resource-efficient model (student) for efficient deployment~\cite{hinton_15}. It enables the student model $M_t$ to learn feature representations from the teacher model $M_{t-1}$ and minimizes divergence between them. KD was first utilized by Li et al.~\cite{lwf} in continual learning within a multi-task framework, where knowledge from a previously trained model is distilled into the current model. Since then, numerous approaches~\cite{lwf, lwm,obj_incr,podnet,lucir} have been developed into two main groups: soft logits-based KD~\cite{lwf, lwm} and feature-based KD~\cite{obj_incr, fitnets, rethink, att_transfer}. Within former techniques, soft logits from the teacher model are distilled to the student model to minimize the KL divergence between their logit distributions. In addition, various other metrics such as a modified cross-entropy~\cite{lwf}, $L_2$-distance~\cite{obj_incr},  Gramian matrix~\cite{gift}, \etc have been designed to reduce the discrepancies between teacher and student models. LwM~\cite{lwm} introduces an attention distillation loss that penalizes the loss of information in the attention maps of the classifiers, guiding the student model to focus on important regions of the input image. FitNets~\cite{fitnets} supervises the student model by minimizing the $L_2$ distance between the intermediate feature maps of the teacher and student networks.

Douillard et al.~\cite{podnet} employ a pooled output distillation (POD) mechanism, which pools and distills the outputs of intermediate feature maps across various stages of the model including the FC layer, to enhance the transfer of knowledge from prior tasks to new ones. Their distillation function minimizes discrepancies of $L_2$-normalized pooled feature maps of input $\textbf{x}$ by utilizing the Euclidean distance, $\mid\mid f^{old}_i (\textbf{x}) - f^{new}_i (\textbf{x})\mid\mid^2_2$. This distillation technique is proven to be effective in mitigating CF over a large number of incremental steps. However, they apply strict constraints on class embeddings, potentially limiting the plasticity of the model. In LUCIR~\cite{lucir}, the authors introduce a distillation loss that focuses solely on the orientation of the normalized feature embeddings. By maximizing the similarity between geometric structures, their approach flexibly retains the spatial configuration of new embeddings to match that of old embeddings.

\section{Proposed Method} \label{sec:method}

\subsection{Problem formulation} 
In our protocol, we split a dataset $\mathcal{D}$ into $\{\mathcal{D}_{train}, \mathcal{D}_{test}\}$, where we train an FR model $M$ using $\mathcal{D}_{train}$ and test it on $\mathcal{D}_{test}$ with no overlaps between identities. The model $M$ is sequentially trained using, $\mathcal{D}_{train} = \{\mathbf{D}^{(t)}\}_{t=1}^{T}$, where $\mathbf{D}^{(t)}$ represents the training data for the $t^{th}$ incremental step, and $T$ is the total number of steps. Furthermore, $\textbf{D}^{(t)}$ comprises $N$ number of samples, ${\{{x}_i^{(t)}, y_i^{(t)}\}}_{i=1}^{N}$, where $x_i^{(t)}$ denotes an image of identity $y_i^t \in {\textbf{C}^{(t)}}$. Note that identities in different steps may overlap, i.e., ${\{\textbf{C}^t \cap \textbf{C}^{t'}\}} \neq \varnothing$ for $t \neq {t'}$. 

In this study, our goal is to incrementally train an FR model over $T$ tasks to effectively identify any unseen face images. After each incremental step $T=t$, we test the latest model $M_t$ twice to determine both in-domain and out-of-domain performance. To evaluate in-domain performance, we test $M_t$ on $\mathcal{D}_{test}$, which has a fixed set of identities $\textbf{C}^{test}$, such that  ${\{\textbf{C}^t \cap \textbf{C}^{test}\}} = \varnothing$ for $t \in {1, \cdots, T}$. To evaluate out-of-domain performance, the model is tested across various FR benchmarks with varying image qualities.

\begin{figure*}[t] 
	\centering
	\includegraphics[width=0.88\linewidth]{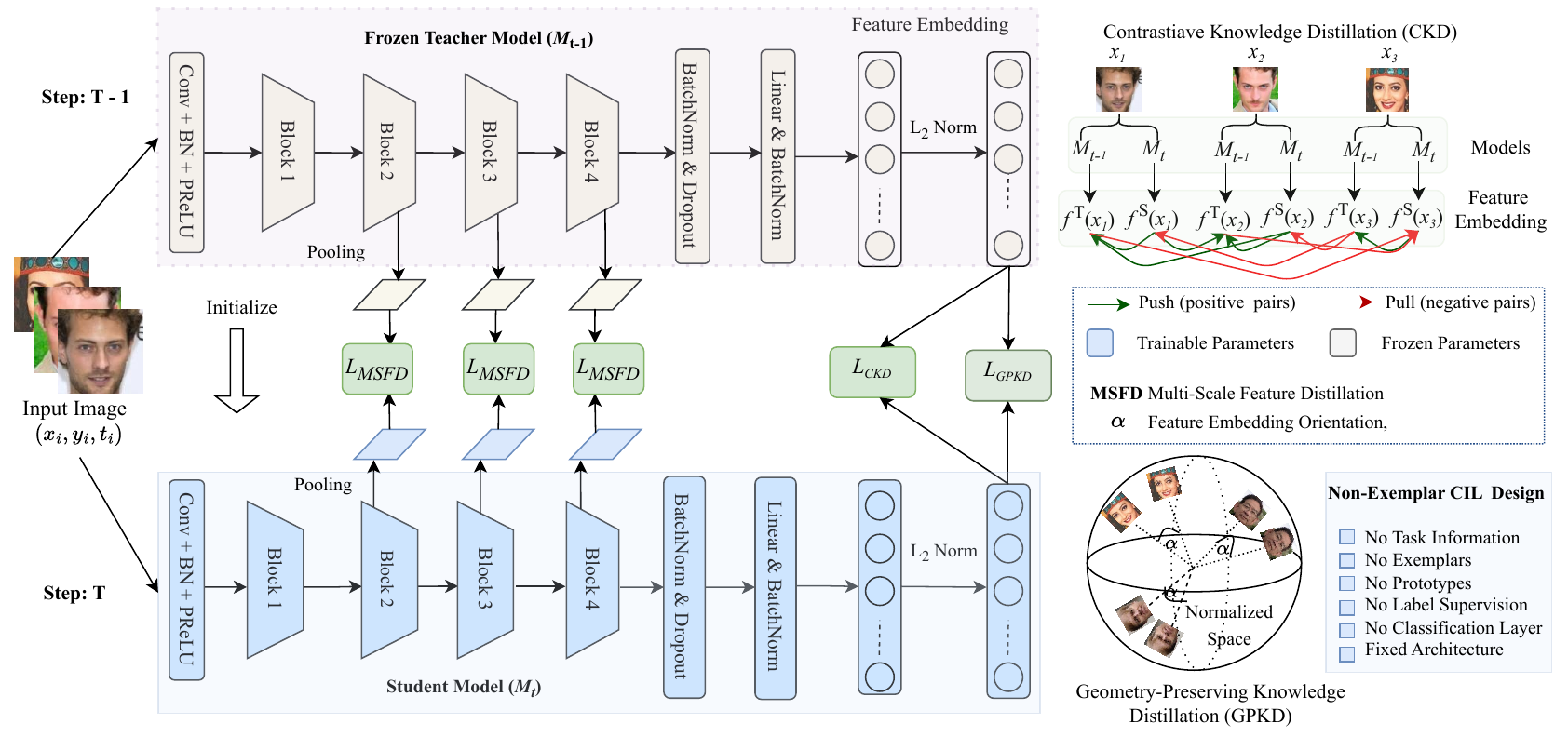}
	\caption{Overview of the proposed CLFace framework for scalable, resource-efficient lifelong face recognition. It consists of a teacher model $M_{t-1}$ trained on previously seen identities, and a student model $M_t$ initialized from $M_{t-1}$. When a new face identity ($x_{i}^{t}$) is fed to both models, the MSFD loss penalizes the drift in local features between the student and teacher models, helping to retain learned feature representations. GPKD constrains the student model to maintain a similar orientation to the teacher model's embedding space, while CKD enhances the discriminative power of the feature representation through similarity matching between new identities.}
	\label{fig:framework}
	\vspace{-2mm}
\end{figure*}

\subsection{Learning paradigm} 
First, we pre-train the base model $M_0$ offline using the base data from each dataset with ArcFace loss~\cite{arcface}. The remaining data are then introduced incrementally in subsequent steps. The proposed CLFace framework, as depicted in Figure~\ref{fig:framework}, includes a teacher model $M_{t-1}$, trained on old data, and a student model $M_t$ initialized from $M_{t-1}$. Our goal is to train $M_t$ to continually learn new identities while preserving the previously learned knowledge. This poses a considerable challenge to the student model $M_t$ as it is presented with an entirely new set of identities. To address this challenge, we design an objective function that includes various distillation losses~\cite{lucir, lwf, lwm}, as outlined in Algorithm~\ref{alg:alg_cl2face}, to penalize the divergence between $M_{t-1}$ and $M_t$. The goal of this objective is to ensure that $M_t$ remains as proficient as $M_{t-1}$ in recognizing the old faces while improving its feature representation to recognize unseen faces effectively. Once $M_t$ is trained for the current step, it becomes the teacher for the next incremental step.

\subsection{Multiscale feature distillation}
Drawing inspiration from prior research~\cite{lin2022knowledge, rethink, fitnets, podnet, att_transfer}, we apply feature-level knowledge distillation to the local feature maps between student and teacher models in our CLFace framework. Traditional CL algorithms typically transfer knowledge at a single-scale feature layer~\cite{lwm, lwf, lucir}, which can limit the student model's ability to fully learn from the teacher model. In contrast, intermediate feature maps at different stages capture various aspects of the input image: lower scales capture fine-grained details, while higher scales capture more coarse-grained information. Hence, we employ a multiscale feature distillation (MSFD) loss to transfer the multi-grained information to the student model, enabling it to learn more comprehensive knowledge from the teacher model.

Let us consider the output of an intermediate stage of the feature extractor, denoted as $f \in \mathcal{R}^{C \times H \times W}$, which consists of $C$ feature planes, each with dimensions $H \times W$. A mapping function $\mathcal{F}$ for this stage processes the 3D feature maps to generate a flattened 2D spatial attention map as follows:
\begin{equation}
	\footnotesize
	\mathcal{F} : R^{~C \times H \times W} \rightarrow R^{~H\times W},
\end{equation}
where $\mathcal{F}$ is the channel-wise pooling operation that computes the spatial attention map. We minimize the $L_2$ norm distance between the corresponding spatial attention maps of the current and the previous models at multiple stages, excluding the first one. This approach provides a more flexible form of knowledge transfer, as applying Euclidean distance directly to the representations of the old model may impose overly rigid constraints on the student model~\cite{podnet}, potentially negatively impacting performance~\cite{lin2022knowledge}. As a result, our MSFD loss, as detailed in Eq.~\ref{eqn:fr-msd}, preserves essential information from learned feature representations while facilitating the learning of new representations:
\begin{equation} 
\label{eqn:fr-msd}
\footnotesize
	\mathcal{L}_{MSFD}=\frac{1}{N(L-1)}\Sigma_{i=1}^N\Sigma_{l=2}^L\mid\mid\phi({\textbf{f}}^{~t-1}_{i, l})-\phi({\textbf{f}}^{~t}_{i, l})\mid\mid_{2}^{2},
\end{equation}
where $\phi$ denotes the channel-wise average pooling operation followed by $L_2$ normalization. $L$ represents the number of stages, and $N$ is the total number of samples. 

\subsection{Geometry-preserving knowledge distillation}
We also introduce a lighter constraint on feature embeddings to achieve an optimal balance between \textit{plasticity} and \textit{rigidity} within our CLFace framework. Building on prior works~\cite{lucir}, this study specifically focuses on local geometric structures, particularly the orientation between normalized feature embeddings. In particular, as illustrated in Figure~\ref{fig:framework}, we design a geometry-preserving knowledge distillation (GPKD) to maintain the orientation of the teacher model's feature embeddings in the student model. GPKD imposes a constraint that prevents the feature embeddings from being entirely rotated. This constraint is crucial for FR, as it computes the similarity between feature embeddings during inference:
\begin{equation}
	\footnotesize
	\mathcal{L}_{GPKD}=\frac{1}{N}\Sigma_{i=1}^N [1 - \langle\frac{\textbf{f}^{~t-1}_{i}}{\mid\mid \textbf{f}^{~t-1}_{i}\mid\mid_2}~,~\frac{\textbf{f}^{~t}_{i}}{\mid\mid\textbf{f}^{~t}_{i} \mid\mid_2}\rangle],
\end{equation}
where $\langle~\rangle$ denotes the cosine similarity between two embedding vectors. Therefore, by emphasizing feature orientation over magnitude, this loss improves the model's adaptability to new identities.

\input{alogrithm/alg}
\subsection{Contrastive knowledge distillation}
The previous two objectives help the student model to retain learned representations during incremental learning. However, we can further leverage the batches of new identities to learn discriminative feature representations. Although label-supervised identity (ID) loss can be used to learn discriminative features, it becomes impractical in lifelong learning scenarios. Alternatively, contrastive knowledge distillation (CKD) allows us to learn discriminative features by maximizing the mutual information between the teacher and student models. Similar to the InfoNCE~\cite{info_nce} loss, we optimize the following contrastive loss, denoted as $L_{CKD}$, as follows:
\begin{equation}
	\footnotesize
	\label{eqn:ckd}
	L_{CKD} =  \frac{-1}{B} \sum_{i=1}^{B} \log \frac{\exp(\langle {\mathbf{\bar{f}}}_i^{~t}, \mathbf{\bar{f}}_i^{~t-1} \rangle / \tau) }{ \sum^{B}_{k=1} \exp (\langle \mathbf{\bar{f}}_i^{~t}, \mathbf{\bar{f}}_k^{~t-1} \rangle / \tau)},
\end{equation}
where $\mathbf{\bar{f}}$ represents the normalized features, $\tau$ denotes the temperature parameter, and $B$ is the batch size. In each batch, the positive pair comprises the feature embeddings from both the student and teacher models for the same identity. All other pairs within the same batch serve as negative pairs since each batch contains unique identities. CKD thus improves discriminative features through a self-supervised learning approach by increasing similarity among positive pairs and decreasing it among negative pairs.

\subsection{Training objective} The total loss of CLFace, as shown in Eq.~\ref{equ:obj_function}, is a weighted summation of these losses:  
\begin{equation}~\label{equ:obj_function}
\displaystyle
L_{total}= \lambda_{1} L_{MSFD} + \lambda_{2} L_{GPKD} + \lambda_{3} L_{CKD}, 
\end{equation}
where $\lambda_{1}$, $\lambda_{2}$, and $\lambda_{3}$ are the weights to balance the losses. 

\vspace*{-1mm}
\section{Experimental Results} \label{sec:exp_results}
\subsection{Datasets and baselines}
\begin{figure*}[t] 
	\centering
	\includegraphics[width=0.80\linewidth]{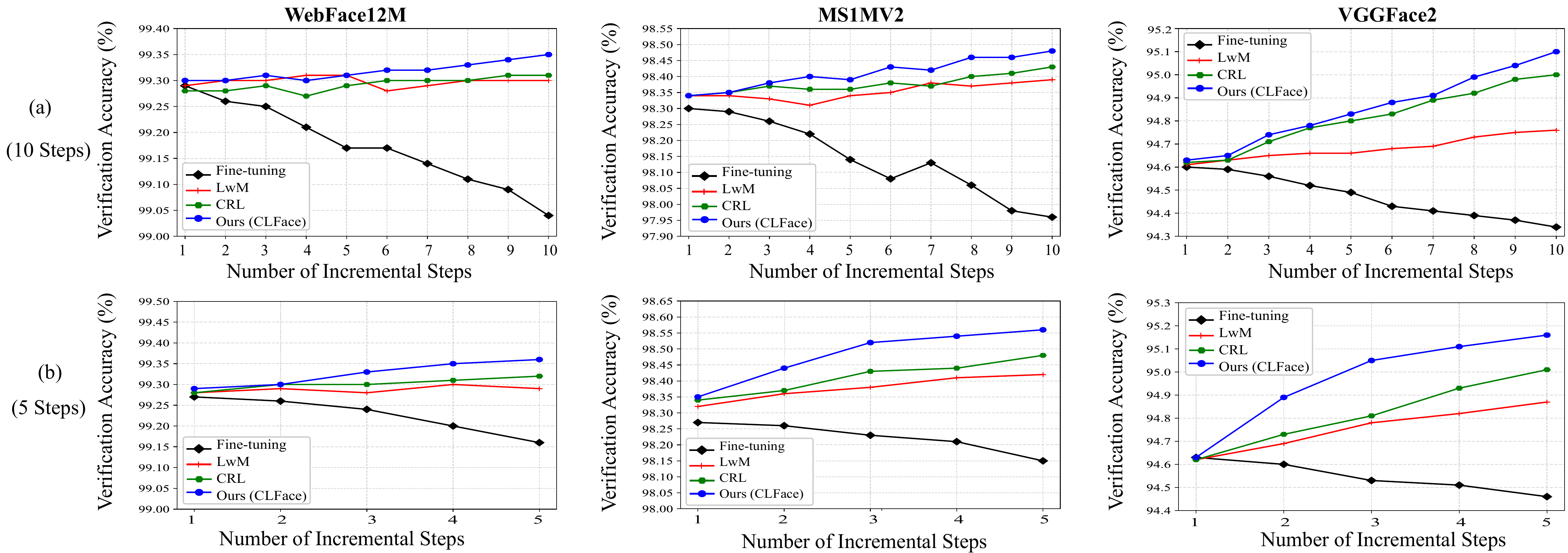}
	\caption{\footnotesize Comparison of CLFace with other approaches for evaluating in-domain performance in (a) 10-step and (b) 5-step learning scenarios.}
	\label{fig:in_domain_eval}
	\vspace{-2mm}
\end{figure*}

State-of-the-art continual learning algorithms typically use small-scale datasets, such as CIFAR-10~\cite{cifar}, CIFAR-100~\cite{cifar}, CUB~\cite{cub}, and ImageNet-1K~\cite{imagenet}, for training and evaluation, which limits their scalability and generalizability to large-scale data. In contrast, deep learning-based FR algorithms require large datasets for training. Additionally, we need to evaluate our CLFace algorithm for both in-domain and out-of-domain performance using large-scale incremental data. In this work, we train our CLFace algorithm using three FR datasets of varying sizes for in-domain performance evaluation. First, we use the small-sized VGGFace2~\cite{vgg_face}, which contains 3.31M images from 9,131 identities. We train CLFace on images from 8,631 identities and evaluate its in-domain performance on a separate test set of 500 identities. The MS1MV2 dataset~\cite{arcface} comprises 5.78M images representing 85,738 identities, making it a medium-sized dataset for training FR algorithms. We train on 85,000 identities and reserve 738 identities for evaluation. We also experiment with the million-scale WebFace12M dataset, a subset of WebFace260M~\cite{wf12m}, which contains over 12M images from 617.97K identities. 

To evaluate the out-of-domain performance of CLFace, we conduct experiments across various FR benchmarks with high, mixed, and low-quality images.  For high-quality images, we use LFW~\cite{lfw}, AGEDB~\cite{agedb}, CALFW~\cite{calfw},  CFPFP~\cite{cfpfp}, and CPLFW~\cite{cplfw}. We follow the \textit{unrestricted with labeled outside data} protocol, where features are trained using additional data. We also tested CLFace on the IJB-B~\cite{ijbb} and IJB-C~\cite{ijbc} datasets for mixed-quality images, and on TinyFace~\cite{tinyface} for low-quality images. Furthermore, we consider different training schemes such as \textit{feature extraction}, \textit{fine-tuning}, and \textit{joint training} as baseline approaches. In particular, feature extraction involves training the model only on the base identities and then extracting features from new identities in subsequent steps. Fine-tuning involves updating the old model with new data, which leads to the forgetting of past knowledge and results in \emph{lower bound} performance. Conversely, joint training uses combined data from all learning steps, providing the \emph{upper bound} performance for any CL algorithm.
\begin{table}[t]
	\centering
	\footnotesize
	\setlength{\tabcolsep}{4pt}
	\caption{Search space and optimal values of the hyperparameters used in the proposed CLFace framework.}
	\begin{tabular}{c cc}
		\toprule[1.0pt]
		Hyperparameter      &Search Space         &Optimum value\\\hline\addlinespace[1mm]
		$\lambda_{1}$    &[0.5 - 5]          &3 \\
		$\lambda_{2}$     &[1- 20]           &12  \\
		$\lambda_{3}$     &[0.1 - 2.0]          &1 \\ 
		$\tau$     &[1.5 - 4.0]         &2  \\
		\bottomrule[1.0pt]
	\end{tabular}
	\label{table:hyperparameter}
	\vspace{-6mm}
\end{table}

\subsection{Implementation Details}
Following standard practice~\cite{podnet, lucir, zhu2021prototype}, half of the identities are reserved for base training, while the remaining identities are randomly and equally divided into either 5 or 10 disjoint sets. The number of images per step varies due to unequal image counts per identity in these datasets. Other approaches~\cite{lwf, lwm, podnet, crl} are also implemented with the same protocol and FR model to ensure a fair comparison. Each dataset is preprocessed by cropping and aligning face images using five landmarks~\cite{arcface}, resulting in $112 \times 112$ images. We utilize the ArcFace~\cite{arcface} FR model with an iResNet50~\cite{resnet, arcface} backbone, setting the scale parameter $s=64$ and the margin parameter $m=0.5$. During each incremental step, CLFace is trained for 10 epochs with an SGD optimizer (mini-batches of 256, weight decay of 0.0005, momentum of 0.9). Here, the initial learning rate is set to 0.01 and is reduced by an exponential decay. Furthermore, the base model is trained for 20 epochs with only ArcFace loss~\cite{arcface}, using the same optimizer settings and starting with a learning rate of 0.1, reduced by a factor of 10 after the 6th and 12th epochs. 

Table~\ref{table:hyperparameter} lists the hyperparameters used in the objective function (see Eq.~\ref{equ:obj_function}), along with their optimal values and search space ranges. The optimal value for each hyperparameter is determined through a grid search within a manually specified range. Experimental results are evaluated using cosine similarity on extracted feature embeddings, with a 10-fold cross-validation scheme. Nine folds are used to determine the threshold for validation, while the 10th fold serves for testing. Results are reported using verification accuracy (VA) from the model trained in the final incremental step (5th or 10th). To account for the potential impact of class sequence on performance, we run our experiments using three random class orders and report the mean results.

\begin{table*}[t]
	\centering
	\setlength{\tabcolsep}{2.1pt} 
	\caption{Comparison of CLFace with other baselines and SoTA methods for out-of-domain evaluation, using the MS1MV2 dataset. Here, * denotes the non-exemplar model (memory budget 0). $S$ denotes total training steps, and $D^t$ represents training data used in $t$-th step.} 
	\label{table:comp_out_of_domain}
	\scriptsize
	\begin{tabular}{c| c ||cc cc cc cc cc|| cc cc|| cc}
		\toprule[1pt]
		&&\multicolumn{10}{c||}{High Quality, VA (\%)}	&\multicolumn{4}{c||}{Mixed Q. (TAR@FAR=0.1\%)} &\multicolumn{2}{c}{Low Q. (Rank-5)}	\\\cline{3-18}
		
		Methods &Step &\multicolumn{2}{c}{LFW~\cite{lfw}}  &\multicolumn{2}{c}{AGEDB~\cite{agedb}} &\multicolumn{2}{c}{CALFW~\cite{calfw}} &\multicolumn{2}{c}{CFPFP~\cite{cfpfp}} &\multicolumn{2}{c||}{CPLFW~\cite{cplfw}} &\multicolumn{2}{c}{IJB-B~\cite{ijbb}} &\multicolumn{2}{c||}{IJB-C~\cite{ijbc}} &\multicolumn{2}{c}{TinyFace~\cite{tinyface}}\\\cline{3-18}
		
		&Data &$S=5$ &$S=10$ &$S=5$ &$S=10$   	&$S=5$ &$S=10$     &$S=5$ &$S=10$ &$S=5$ &$S=10$  &$S=5$ &$S=10$ &$S=5$ &$S=10$   &$S=5$ &$S=10$  \\\midrule 
		Feature Ex.      &0\%   &99.74 &99.74  &97.01 &97.01  &95.72 &95.72     &94.62 &94.62  &89.54 &89.54  &91.0  &91.0 &92.94 &92.94 &62.47 &62.47\\ 
				
		Fine-tuning      &$D^t$   &99.58 &99.55  &95.65 &94.77  &95.13 &94.82     &94.53 &94.13  &88.90 &88.63  &90.90  &87.63 &92.72 &90.05 &61.21 &59.82 \\ 

		Joint Train.   &100\% &\textbf{99.80} &\textbf{99.80}  &\textbf{98.08} &\textbf{98.08}  &\textbf{96.05} &\textbf{96.05}  &\textbf{97.80} &\textbf{97.80}  &\textbf{92.15} &\textbf{92.15}  &\textbf{95.0}  &\textbf{95.0} &\textbf{96.46} &\textbf{96.46} &\textbf{67.54} &\textbf{67.54}\\
		LWF-MC   &\multirow{5}{*}{$D^t$} &99.61 &99.58  &95.83 &95.22  &95.20 &95.06     &94.56 &94.19  &88.98 &88.75  &-     &- &- &- &- &-\\
		LwM   	 & &99.66 &99.62  &96.15 &95.78  &95.23 &95.09     &94.58 &94.21  &89.05 &88.96  &-     &- &- &- &- &-\\ 
		PODNet*  & &99.70 &99.64  &95.84 &95.81 	&95.21 &95.13     &94.62 &94.46  &89.11 &89.0   &-     &- &- &- &- &- \\ 
		CRL   	 & &99.75 &99.72  &96.94 &96.85  &95.73 &95.68     &94.65 &94.57  &89.60 &89.48  &-     &- &- &- &- &-\\
		
		\textbf{CLFace}      & &99.75 &99.75 &97.05 &97.02   &95.77 &95.77     &94.72 &94.71  &89.95 &89.70  &91.37 &91.04 &93.32 &93.05 &64.03 &62.96 \\ 
		\bottomrule[1pt]
	\end{tabular}
\vspace{-3mm}
\end{table*}
\subsection{Evaluation on in-domain datasets}
We compare the proposed CLFace with SoTA approaches like LwM~\cite{lwm} and CRL~\cite{crl}, along with the baseline fine-tuning approach, to evaluate in-domain performance on the WebFace12M~\cite{wf12m}, MS1MV2~\cite{arcface}, and VGGFace2~\cite{vgg_face} datasets across both 5-step and 10-step incremental learning scenarios. We select approaches like LwM~\cite{lwm} and CRL~\cite{crl} because they can be implemented using our protocol, whereas other methods~\cite{icarl, ewc, lucir} are not scalable for such protocols. Figure~\ref{fig:in_domain_eval} illustrates that the proposed CLFace achieves the highest VA(\%) across all three datasets, whereas the fine-tuning scheme performs worse in both learning scenarios. Specifically, CLFace shows VA improvements over CRL~\cite{crl} of 0.04\%, 0.05\%, and 0.10\% for 10-step learning, and 0.04\%, 0.05\%, and 0.10\% for 5-step learning, across these datasets, respectively. Notably, the performance improvement on VGGFace2 is higher than on WebFace12M during incremental steps. This discrepancy arises because the larger base data in WebFace12M allows the model to be efficiently trained from scratch. Therefore, the additional incremental data contributes less significantly in improving feature representations in WebFace12M compared to VGGFace2.

\subsection{Evaluation on out-of-domain datasets}
Furthermore, we evaluate CLFace's out-of-domain performance across various FR benchmarks with high, mixed, and low-quality images. We specifically tested CLFace on high-quality benchmarks including LFW~\cite{lfw}, AGEDB~\cite{agedb}, CALFW~\cite{calfw},  CFPFP~\cite{cfpfp}, and CPLFW~\cite{cplfw} on VA. As shown in Table~\ref{table:comp_out_of_domain}, our method consistently outperforms baseline approaches and SoTA models~\cite{lwf, lwm, podnet, crl} across all datasets. As expected, the fine-tuning scheme results in the lowest VA, while the joint training scheme achieves the highest VA due to its standard multitask setup rather than incremental training. Compared to the second-best performing method, CRL~\cite{crl}, our CLFace demonstrates improvements in VA of 0.03\%, 0.17\%, 0.09\%, 0.14\%, and 0.22\% in 10-step learning, and 0.0\%, 0.11\%, 0.02\%, 0.07\%, and 0.35\% in 5-step learning scenarios across these datasets, respectively. However, the performance in 10-step learning is lower than in 5-step learning scenarios, suggesting that performance improvement is higher when incremental steps include data from a large number of identities, irrespective of old or new.

Additionally, we evaluate CLFace on the challenging IJB-B~\cite{ijbb} and IJB-C~\cite{ijbc} benchmarks, which include both high and low-quality images, using a 1:1 verification scheme with the TAR@FAR=0.0001 metric. We also test CLFace on the TinyFace dataset~\cite{tinyface} for performance on low-quality, in-the-wild images using the Rank-5 metric. Despite a widening performance gap compared to the upper bound, CLFace outperforms baseline approaches, demonstrating its generalizability in real-world settings.

\subsection{Ablation study}
We evaluate CLFace through ablation studies on various components of the framework. All experiments are performed on the MS1MV2~\cite{arcface} dataset with a 5-step incremental learning scenario, using the VA (\%) metric.
\vspace*{-3mm}
\paragraph{Analysis on objective function:\hspace*{-2mm}}
\begin{table}[t]
	\centering
	\footnotesize
	\setlength{\tabcolsep}{2.5pt}
	\caption{An ablation study on the effectiveness of different losses in the CLFace framework}
	
	\begin{tabular}{ccc| ccccc }
		\toprule[1pt]
		MSFD &GPKD &CKD 			&LFW &AGEDB  &CALFW &CFPFP  &CPLFW  \\\midrule
		
		\cmark &\xmark 	&\xmark 	&99.55 &95.67 &95.06 &92.71  &89.03 \\
		\xmark &\cmark 	&\xmark 	&99.70 &97.0  &95.57 &94.19  &89.42 \\
		\xmark &\xmark 	&\cmark 	&99.74 &96.88 &95.44 &94.07  &89.30\\
		\cmark &\cmark  &\xmark 	&99.71 &96.85 &95.61 &94.34  &89.58 \\
		\cmark &\cmark  &\cmark 	&\textbf{99.75} &\textbf{97.05} &\textbf{95.77} &\textbf{94.72}  &\textbf{89.95} \\
		\bottomrule[1pt]
	\end{tabular}
	\label{table:abs_obj_function}
	\vspace{-2mm}
\end{table}
We conduct an ablation study to analyze the impact of each loss function in the CLFace framework. As shown in Table~\ref{table:abs_obj_function}, each loss function contributes to overall performance, with the final objective function achieving the highest VA(\%). Although MSD is less effective than GPKD in retaining representations, the combination of MSD and GPKD outperforms CKD alone. The final objective achieves improvements of 0.04\%, 0.20\%, 0.16\%, 0.38\%, and 0.37\% over MSD + GPKD on LFW~\cite{lfw}, AGEDB~\cite{agedb}, CALFW~\cite{calfw},  CFPFP~\cite{cfpfp}, and CPLFW~\cite{cplfw}, respectively. This proves the complementary nature of these loss functions.
\vspace{-2mm}
\paragraph{Analysis on base training:\hspace{-2mm}}
\begin{table}[t]
	\centering
	\setlength{\tabcolsep}{4pt} 
	\caption{An ablation study to examine the impact on varying amounts of base data on verification performance} \label{table:abs_base_training}
	\scriptsize
	\begin{tabular}{ccc cc cccc}
		\toprule [1pt]
		\multirow{2}{*}{Base} &\multirow{2}{*}{Methods} &\multirow{2}{*}{Metric} &\multicolumn{6}{c}{Steps} \\\cline{4-9}
		
		&  &&base	&1	&2	&3 &4	&5         \\ \hline
		\multirow{5}{*}{10\%} &CALFW  &\multirow{3}{*}{VA(\%)}	&93.10 &93.23 &93.30 &93.30   	&93.32 &93.35 \\
							  &CFPFP	  &&91.87 &91.95 &92.07 &92.17   	&92.21 &92.29 \\ 
							  &CPLFW	  & &86.30 &86.50 &86.61 &86.78   	&86.75 &86.81 \\ 
							  &IJB-C	  &TAR &87.63 &87.65 &87.72 &87.82 &87.92 &87.89 \\
							  &TinyFace   &Rank-5 &55.69 &55.90 &56.16 &56.22 &56.41  &57.0\\ \hline\rule{0pt}{2mm}

		\multirow{5}{*}{25\%} &CALFW  &\multirow{3}{*}{VA(\%)} &94.96 &94.96 &94.98 &95.03   	&95.03 &95.03 \\ 
							  &CFPFP			&&91.80 &92.10 &92.25 &92.34   	&92.27 &92.34 \\
							  &CPLFW			&&87.95 &88.0  &88.06 &88.03   	&88.10 &88.18 \\ 
							  &IJB-C	 &TAR &89.69 &89.66 &90.24 &89.75 &89.77 &89.99 \\
							  &TinyFace  &Rank-5 &59.03 &59.03 &59.50 &59.50	&59.38 &59.62 \\ \hline\rule{0pt}{2mm}

		\multirow{5}{*}{50\%} &CALFW  &\multirow{3}{*}{VA(\%)} &95.72 &95.74 &95.75 &95.75   	&95.76 &95.77 \\ 
		 					  &CFPFP			&&94.62 &94.67 &94.70 &94.68   	&94.70 &94.72\\ 
							  &CPLFW     		&&89.54 &89.55 &89.61 &89.69   	&89.78 &89.95\\
							  &IJB-C	 &TAR  &92.94 &93.02 &93.24 &93.36 &93.46 &93.32 \\
							  &TinyFace  &Rank-5 &62.47 &63.95 &63.95 &64.04		&63.46 &64.03 \\ \hline\rule{0pt}{2mm}

		\multirow{5}{*}{75\%} &CALFW &\multirow{3}{*}{VA(\%)} 		&95.88 &95.93 &95.93 &\textbf{95.96}   	&95.95 &95.95 \\ 
							  &CFPFP			&&96.19 &96.28 &96.34 &96.37   	&96.41 &\textbf{96.42} \\
							  &CPLFW			&&91.20 &91.18 &91.20 &91.21   	&91.21 &\textbf{91.22} \\ 
							  &IJB-C	 &TAR &94.55 &94.62 &\textbf{94.80} &94.62 &95.57 &94.68 \\
							  &TinyFace  &Rank-5 &64.48 &64.58 &64.80 &64.91		&64.94 &\textbf{64.99} \\
		\bottomrule [1pt]
	\end{tabular}
	\vspace{-4mm}
\end{table}
Next, we investigate the effect of base training on the CLFace framework, as detailed in Table~\ref{table:abs_base_training}. Our analysis, conducted on five benchmark FR datasets of varying qualities, yields several key observations: (I) The performance of CLFace significantly improves with the increase in the base data. For example, using 75\% base data improves performance more than using just 25\% base data across these datasets. (II) The performance of CLFace is improved with incremental updates, though the improvement is inconsistent. This observation confirms that CLFace effectively mitigates CF compared to other methods. (III) The degree of improvement over incremental steps heavily depends on the proportion of training data used in each step. For example, when incremental data makes up 90\% of the training set (1st row), improvements of 0.25\%, 0.42\%, and 0.51\% in VA(\%), 0.26\% in TAR@FAR=0.1\%, and 1.31\% in Rank-1 are observed on CALFW, CFPFP, CPLFW, IJB-C, and TinyFace, respectively. This is higher than when incremental data constitutes 25\% of the training data (4th row). These results highlight the importance of both base training and the amount of incremental data in optimizing CLFace performance.
\vspace{-3mm}
\paragraph{Analysis on identity supervision:\hspace*{-2mm}} We further carry out an ablation study to examine the effect of removing label supervision from the CLFace framework. Table~\ref{table:abs_identity} reports VA(\%) results across various datasets demonstrating that CLFace performs better when ID loss is combined with GPKD loss. In contrast, ID+CKD losses degrade CLFace performance. Both losses focus on learning discriminative features, which induces a bias toward new identities and eventually leads to CF. The combination of all losses yields moderate performance, it remains below that of the ID + GPKD combination. CKD loss, however, is more compatible with lifelong learning compared to ID loss. Furthermore, our objective, MFSD + GPKD + CKD, shows improvements of 0.53\% on AGEDB and 0.14\% on CALFW, but also degradations of 1.06\% on CFPFP and 0.11\% on CPLFW compared to the best-performing objective with ID loss (ID + GPKD). These results suggest that replacing ID supervision with CKD loss does not compromise the generalizability of the CLFace model, while also benefiting from label-free supervision in lifelong learning.

\begin{table}[t]
	\centering
	\footnotesize
	\setlength{\tabcolsep}{2.5pt}
	\caption{An ablation study to evaluate the impact of removing identity supervision on performance with unseen identities}
	
	\begin{tabular}{cccc| ccccc }
		\toprule[1pt]
		ID &MSFD &GPKD &CKD 		        &LFW &AGEDB  &CALFW &CFPFP  &CPLFW  \\\midrule
		
		\cmark &\xmark &\xmark 	&\xmark 	&99.58 &95.65 &95.13 &94.53  &88.90\\
		\cmark &\cmark &\xmark 	&\xmark 	&99.70 &95.78 &95.09 &94.62  &89.03 \\
		\cmark &\xmark &\cmark 	&\xmark 	&\textbf{99.75} &96.52 &\textbf{95.63 }&95.78  &\textbf{90.06} \\ 
		\cmark &\xmark &\xmark 	&\cmark 	&99.58 &95.32 &94.80 &93.85  &88.70\\ 
		\cmark &\cmark &\cmark  &\cmark 	&99.70 &\textbf{96.85} &95.55 &\textbf{96.35}  &90.05 \\
		\bottomrule[1pt]
	\end{tabular}
	\label{table:abs_identity}
	\vspace{-4mm}
\end{table}

\subsection{Discussion}
In real-world BI systems, traditional CL methods attempt to mitigate CF on seen identities but struggle with unseen ones. Joint training yields better results but requires all training data to be present upfront. In contrast, our CLFace method addresses these challenges by (I) providing a lightweight CL framework that remains consistent during lifelong learning; (II) boasting base performance during incremental steps while outperforming baselines; and (III) leveraging unlabeled face datasets for open-set face recognition. However, CLFace has limitations: its ability to improve feature representations diminishes with more incremental steps, and it performs poorly on low-quality images compared to high-quality ones. Therefore, lifelong face recognition remains an open challenge. Future work could address these issues and extend the approach to other biometric tasks like iris and palmprint recognition.

\section{Conclusion} \label{sec:conclusions}
We present CLFace, a scalable and resource-efficient lifelong face recognition algorithm designed to learn millions of identities from continuous data streams for real-world applications. CLFace uses multiscale and geometry-preserving distillation loss in the feature space to learn and improve feature representations without storing exemplars or requiring label supervision during incremental training. Extensive experiments on unseen identities, across both in-domain and out-of-domain datasets, show that CLFace outperforms state-of-the-art methods and baseline approaches.

\section{Acknowledgment}
This material is based upon a work supported by the Center for Identification Technology Research and the National Science Foundation under Grant 1650474.

{\small
\bibliographystyle{ieee_fullname.bst}
\bibliography{main.bib}
}

\end{document}

%% file: alogrithm/alg.tex
\begin{algorithm}[t]
\footnotesize
\caption{\small{CL2Face Implementation}}
\label{alg:alg_cl2face}
\begin{algorithmic}[1]

\STATE \textbf{Input}: 
Number of steps $T$, \\
Number of epochs per steps $E$, \\
Set of face identities in step $t$, $\mathbf{C}^t$, \\
Number of identities in step $t$ is $N_t$, \\
Datasets, $\mathcal{D}_{train} = \{(x^{(t)}, y^{(t)})\}^T_{t=1}$, \\
Number of batches per epoch, $N_b$; Batch size, $B$.

\STATE \textbf{Require}: Base Model $M_0$
\FOR{$t = 0, 1, \cdots, T$}

\STATE Form step dataset, $\textbf{D}^{(t)} \gets \{(x^t_i, y^t_i)\}^{N}_{i=1}$, where $y^{t}_{i} \in~\mathbf{C}^t$\\

\IF{$t == 0$}
\STATE Randomly initialize the base model ${M_0}$ with $\boldsymbol{\theta}_0$
\STATE Randomly initialize the classification layer ${CL}$ with $\boldsymbol{\theta}_{cl}$
\ELSIF {$t > 0$}
\STATE Initialize, student model ${M_t}$, $\boldsymbol{\theta}_t \gets \boldsymbol{\theta}_{t-1}$ of teacher $M_{t-1}$
\ENDIF

\FOR{$e = 1, \cdots, E$}
\FORALL{$i \in \{1, \cdots , N_b\}$}
\STATE Draw a mini-batch of $\{(x_{i}^{t}, y_{i}^{t})\}_{i=1}^{i=|B|}$ from $\textbf{D}^{(t)}$\\

\IF {$t == 0$}
\STATE $y_{cl_i}^{t} \gets CL(M_{0}(x_{i}^{t},\boldsymbol{\theta}_0), \boldsymbol{\theta}_{cl})$,   \hfill //compute base scores
\STATE $\mathcal{L} \gets \mathcal{L}_{CE} (y_{cl_i}^{t}, y_{i}^{t})$ \hfill //ArcFace loss~\cite{arcface}

\ELSIF {$t > 0$}
\STATE $f_{i}^{t}, f_{2i}^{t}, f_{3i}^{t}, f_{4i}^{t} \gets M_{t}(x_{i}^{t}, \boldsymbol{\theta}_t)$,   \hfill //compute new feat.
\STATE $ f_{i}^{t-1}, f_{2i}^{t-1}, f_{3i}^{t-1}, f_{4i}^{t-1} \gets M_{t-1}(x_{i}^{t}, \boldsymbol{\theta}_{t-1})$,  \hfill //old 
\STATE $\bar{f}_{i}^{t} = L_2 (f_{i}^{t})$;~~~$\bar{f}_{i}^{t-1} = L_2 (f_{i}^{t-1})$ \hfill //$L_2$-Norm
\STATE $\mathcal{L} \gets \lambda_{1} \cdot \mathcal{L}_{MSFD} ((f_{2i}^{t}, f_{3i}^{t}, f_{4i}^{t}),  \newline(f_{2i}^{t-1}, f_{3i}^{t-1}, f_{4i}^{t-1})) + \lambda_{2} \cdot \mathcal{L}_{GPKD}(\bar{f}_{i}^{t}, \bar{f}_{i}^{t-1} ) +  \lambda_{3} \cdot \mathcal{L}_{CKD} (\bar{f}_{i}^{t}, \bar{f}_{i}^{t-1})$ \hfill //by Eq. ~\ref{equ:obj_function}
\ENDIF
\STATE Update  $\boldsymbol{\theta}_t$ //using $\nabla \mathcal{L}$

\ENDFOR
\ENDFOR
\ENDFOR
\end{algorithmic}
\end{algorithm}